\documentclass[10pt,twocolumn,letterpaper]{article}

\usepackage{cvpr}
\usepackage{times}
\usepackage{epsfig}
\usepackage{graphicx}
\usepackage{amsmath}
\usepackage{amssymb}

\usepackage{algorithm}
\usepackage{algorithmic}
\usepackage{booktabs}
\usepackage{subcaption}

\newlength\myindent%
\newcommand\bindent{%
	\begingroup
	\setlength{\itemindent}{\myindent}
	\addtolength{\algorithmicindent}{\myindent}
}
\newcommand\eindent{\endgroup}

\usepackage[breaklinks=true,bookmarks=false]{hyperref}

\cvprfinalcopy % *** Uncomment this line for the final submission

\def\cvprPaperID{3827} % *** Enter the CVPR Paper ID here

\ifcvprfinal\fi

\begin{document}

%%%%%%%%% TITLE
\title{SplineCNN: Fast Geometric Deep Learning with Continuous B-Spline Kernels}

\author{Matthias Fey$^\ast$, Jan Eric Lenssen$^\ast$,  Frank Weichert,  Heinrich M\"uller\\
Department of Computer Graphics\\
TU Dortmund University\\
{\tt\small \{matthias.fey,janeric.lenssen\}@udo.edu}\\
$\small\ast$ \small{Both authors contributed equally to this work.} 
}

\maketitle
\thispagestyle{empty}

\begin{abstract}
	We present \emph{Spline-based Convolutional Neural Networks (SplineCNNs)}, a variant of deep neural networks for irregular structured and geometric input, \eg, graphs or meshes. Our main contribution is a novel convolution operator based on B-splines, that makes the computation time independent from the kernel size due to the local support property of the B-spline basis functions.
	As a result, we obtain a generalization of the traditional CNN convolution operator by using continuous kernel functions parametrized by a fixed number of trainable weights.
	In contrast to related approaches that filter in the spectral domain, the proposed method aggregates features purely in the spatial domain.
	In addition, SplineCNN allows entire end-to-end training of deep architectures, using only the geometric structure as input, instead of handcrafted feature descriptors.
	
	For validation, we apply our method on tasks from the fields of image graph classification, shape correspondence and graph node classification, and show that it outperforms or pars state-of-the-art approaches while being significantly faster and having favorable properties like domain-independence.
	Our source code is available on GitHub\footnote{\url{https://github.com/rusty1s/pytorch_geometric}}.
\end{abstract}

\vspace{-0.6cm}

\section{Introduction}

Most achievements obtained by deep learning methods over the last years heavily rely on properties of the convolution operation in convolutional neural networks~\cite{Lecun1998}: local connectivity, weight sharing and shift invariance.
Since those layers are defined on inputs with a grid-like structure, they are not trivially portable to non-Euclidean domains like discrete manifolds, or (embedded) graphs.
However, a large amount of data in practical tasks naturally comes in the form of such irregular structures, \eg~graph data or meshes.
Transferring the high performance of traditional convolutional neural networks to this kind of data holds the potential for large improvements in several relevant tasks.
\begin{figure}[tb]
	\centering
	\begin{subfigure}[b]{0.45\linewidth}
		\includegraphics[width=1\textwidth]{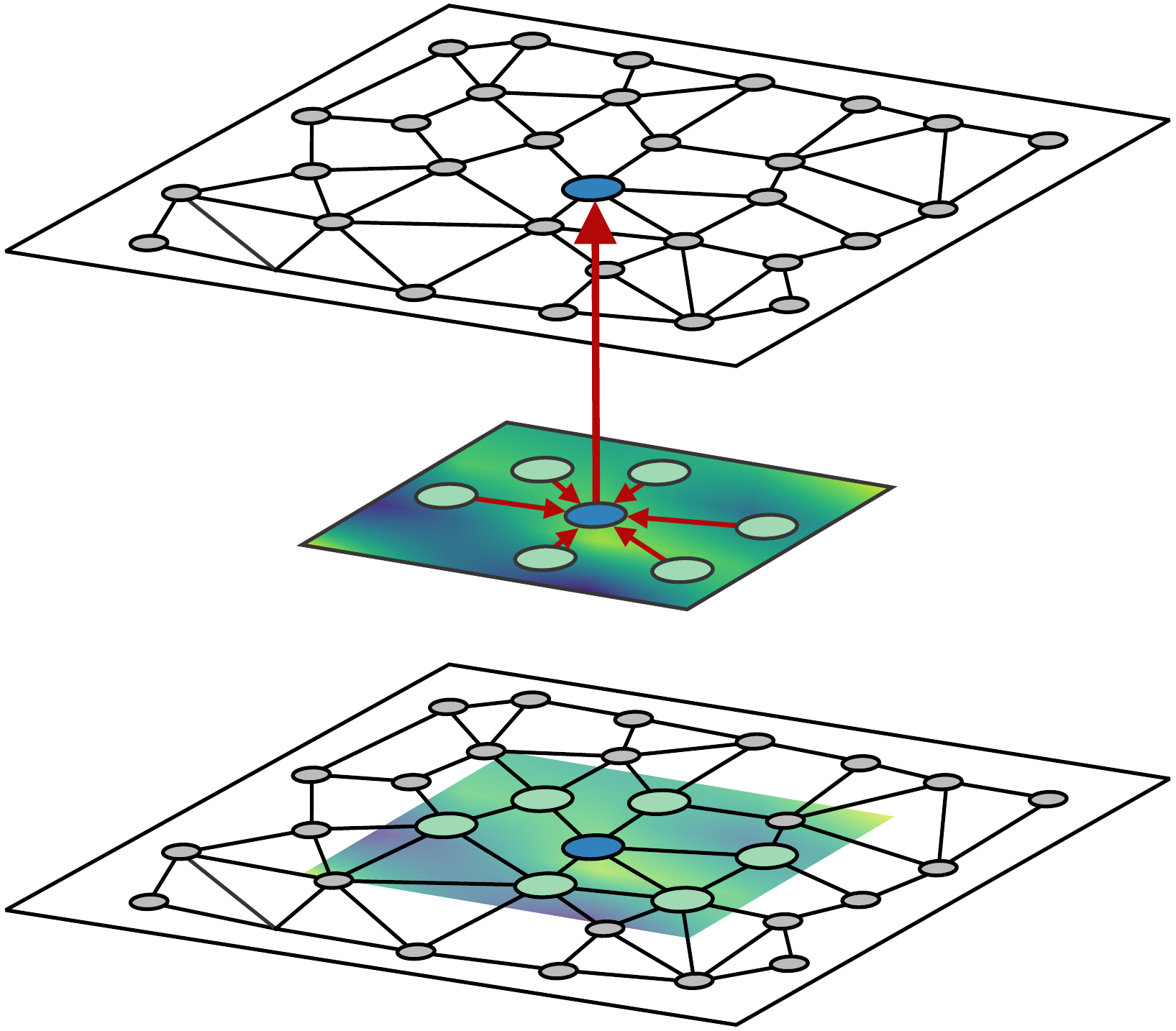}
		\caption{Filtering of graphs}\label{fig:intro1}
	\end{subfigure}
	\hfill
	\begin{subfigure}[b]{0.53\linewidth}
		\includegraphics[width=1\textwidth]{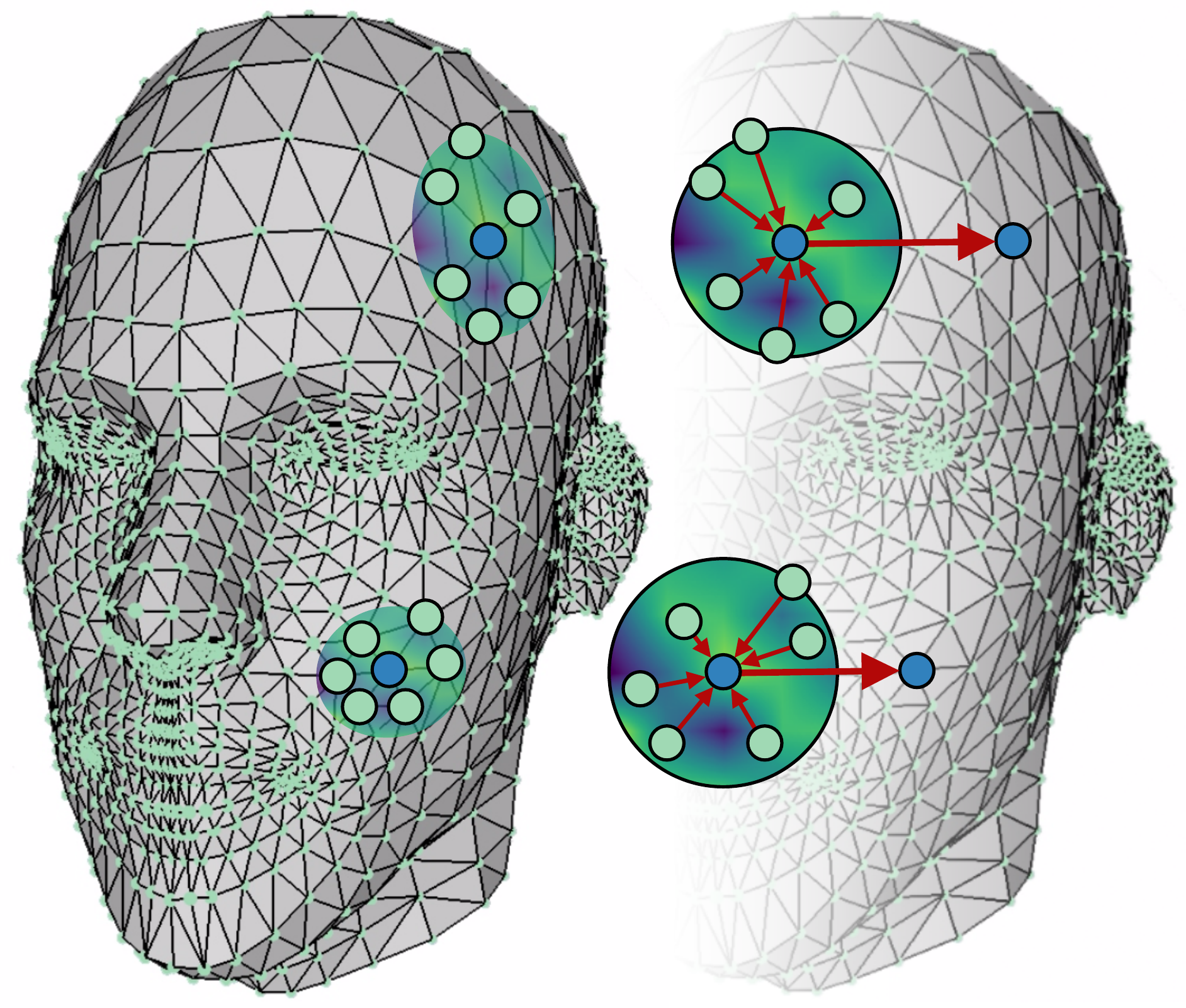}
		\caption{Filtering of meshes}\label{fig:intro2}
	\end{subfigure}
	\caption{Examples for spatial aggregation in geometric deep learning with trainable, continuous kernel functions, showing methods for (a) image graph representations and (b) meshes.}\label{fig:intro_image}
\end{figure}

Recently, a set of methods brought together under the term \emph{geometric deep learning}~\cite{Bronstein2017} emerged, which aim to achieve this transfer by defining convolution operations for deep neural networks that can handle irregular input data.
Existing work in this field can loosely be divided into two different subsets: the spectral and the spatial filtering approaches.
The former is based on spectral graph theory~\cite{Chung1997}, where eigenvalues of a graph's Laplacian matrix are interpreted as frequencies of node signals~\cite{Shuman2013}.
They are filtered in the spectral domain, ana\-logously to Fourier domain filtering of traditional signals.
The latter subset, the spatial approaches, perform convolution in local Euclidean neighborhoods \wrt~local positional relations between points, represented for example as polar, spherical or Cartesian coordinates, as shown as examples in Figure~\ref{fig:intro_image}.

\paragraph{Contribution.}

We present \emph{Spline-based Convolutional Neural Networks (SplineCNNs)}, a variant of deep neural networks for irregular structured data.
The main contribution is a trainable, spatial, continuous convolution kernel that leverages properties of B-spline bases to efficiently filter geometric input of arbitrary dimensionality.
We show that our method
\begin{itemize}
	\setlength\itemsep{0.1em}
	\item can be applied on different kinds of irregular structured data, \eg, arbitrary (embedded) graphs and meshes,
	\item uses spatial geometric relations of the input,
	\item allows for end-to-end training without using handcrafted feature descriptors, and
	\item improves or pars the state-of-the-art in geometric learning tasks.
\end{itemize}
In addition, we provide an efficient GPGPU algorithm and implementation that allows for fast training and inference computation.

\section{Related work}

\paragraph{Deep learning on graphs.}

The history of geometric deep learning began with attempts to generalize convolutional neural networks for graph inputs.
A large number of successful approaches are based on spectral graph theory.
Bruna \etal~\cite{Bruna2014} introduced convolution-like operators on spectral graphs, interpreting the eigenvectors of the Laplacian as Fourier basis.
As an extension, Henaff \etal~\cite{Henaff2015} suggest to use spline interpolation for smoothing kernels in the spectral domain.
Defferrard \etal~\cite{Defferrard2016} approximates spectral filters with Chebyshev polynomials, providing a more efficient filtering algorithm, whose kernel size determines the range of aggregated local $K$-neighborhoods.
This approach was further simplified by Kipf and Welling~\cite{Kipf2017}, who consider only the one-neighborhood for one filter application.
A filter based on the Caley transform was proposed as an alternative for the Chebyshev approximation by Levie \etal~\cite{Levie2017}.
Together with a trainable zooming parameter, this results in a more stable and flexible spectral filter.

It should be noted that all these spectral approaches assume that information is only encoded in the connectivity, edge weights and node features of the input.
While this is true for general graphs, it does not hold for embedded graphs or meshes, where additional information is given by relative positions of nodes, which we consider with our method.

A downside of many spectral approaches is the fact that they use domain-dependent Fourier bases, which restricts generalization to inputs with identical graph connectivity.
Yi \etal~\cite{Yi2017} tackle this problem by applying a spectral transformer network that synchronizes the spectral domains.
Since our approach works directly in the spatial domain, it is not prone to this problem.

For the shape correspondence task on meshes, which we also analyze in this work, Litany \etal~\cite{Litany2017} present a siamese network using a soft error criterion based on geodesic distances between nodes. We compare our method against this specialized method.

\paragraph{Local descriptors for discrete manifolds.}

The issue of not representing local positional relations can be tackled by using methods that extract representations for local Euclidean neighborhoods from discrete manifolds.

Based on the intrinsic shape descriptors of Kokkinos \etal~\cite{Kokkinos2012}, Masci \etal~\cite{Masci2015} present such a method for extraction of two-dimensional Euclidean neighborhoods from meshes and propose a convolution operation locally applied on these neighborhoods.
Boscaini \etal~\cite{Boscaini2016} improve this approach by introducing a patch rotation method to align extracted patches based on the local principal curvatures of the input mesh.

Our convolution operator can but does not have to receive those local representations as inputs.
Therefore, our approach is orthogonal to improvements in this field.

\paragraph{Spatial continuous convolution kernels.}

While the first continuous convolution kernels for graphs work in the \emph{spectral} domain (\eg~\cite{Henaff2015, Defferrard2016, Schutt2017}), \emph{spatial} continuous convolution kernels for irregular structured data were introduced recently as a special case in the fields of neural message passing and self-attention mechanisms~\cite{Gilmer2017, Simonovsky2017, Monti2016}. Furthermore, Monti \etal~\cite{Monti2016} presented the MoNet framework for interpreting different kind of inputs as directed graphs, on which we built upon in our work.
We show that our kernels achieve the same or better accuracy as the trainable Gaussian mixture model (GMM) kernels of MoNet, while being able to be trained directly on the geometric structure.

\section{SplineCNN}\label{sec:methods}

We define SplineCNNs as a class of deep neural networks that are built using a novel type of spline-based convolutional layer. This layer receives irregular structured data, which is mapped to a directed graph, as input.
In the spatial convolutional layer, node features are aggregated using a trainable, continuous kernel function, which we define in this section.

\subsection{Preliminaries}\label{sec:background}

\paragraph{Input graphs.}

Similar to the work of Monti \etal~\cite{Monti2016}, we expect the input of our convolution operator to be a directed graph $\mathcal{G} = (\mathcal{V}, \mathcal{E}, \mathbf{U})$ with $\mathcal{V} = \{1, \ldots, N\}$ being the set of nodes, $\mathcal{E} \subseteq \mathcal{V} \times \mathcal{V}$ the set of edges, and \mbox{$\mathbf{U} \in {[0,1]}^{N\times N \times d}$} containing $d$-dimensional pseudo-coordinates \mbox{$\mathbf{u}(i,j) \in {[0,1]}^d$} for each directed edge \mbox{$(i,j) \in \mathcal{E}$}.
Note that $\mathbf{U}$ can be interpreted as an adjacency matrix with $d$-dimensional, normalized entries \mbox{$\mathbf{u}(i,j) \in {[0,1]}^d$ if $(i,j) \in \mathcal{E}$} and $\mathbf{0}$ otherwise.
Also, $\mathbf{U}$ is usually sparse with $E = |\mathcal{E}| \ll N^2$ entries.
For a node $i \in \mathcal{V}$ its \emph{neighborhood} set is denoted by $\mathcal{N}(i)$.

\paragraph{Input node features.}

Let \mbox{$\mathbf{f}: \mathcal{V} \rightarrow \mathbb{R}^{M_{\textrm{in}}}$}, with \mbox{$\mathbf{f}(i) \in \mathbb{R}^{M_{\textrm{in}}}$}, denote a vector of $M_{\textrm{in}}$ input features for each node $i \in \mathcal{V}$.
For each \mbox{$1 \leq l \leq M_{\textrm{in}}$} we reference the set $\{ f_l(i) \mid i \in \mathcal{V} \}$ as \emph{input feature map}.

\paragraph{B-spline basis functions.}

In addition to the input graph and node features, let $({(N_{1,i}^m)}_{1\leq i \leq k_1}, \ldots, {(N_{d,i}^m)}_{1\leq i \leq k_d})$ denote $d$ open B-spline bases of degree $m$, based on uniform, \ie~equidistant, knot vectors (\cf~Piegl \etal~\cite{Piegl1997}), with $\mathbf{k} = (k_1, \ldots ,k_d)$ defining our $d$-dimensional kernel size.

\begin{figure}[t]
	\includegraphics[width=1.015\linewidth]{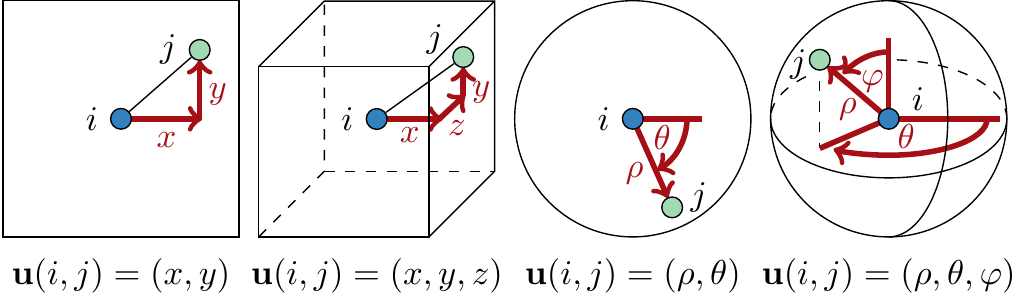}
	\caption{Possibilities for pseudo-coordinates $\mathbf{u}$: two- and three-dimensional Cartesian, polar and spherical coordinates. Values for scaling and translation of the coordinates $\mathbf{u}$ to interval ${[0, 1]}^d$ are omitted.}%
	\label{fig:pseudo}
\end{figure}
\begin{figure*}[t]
	\hspace{0.5cm}
	\begin{subfigure}[t]{0.45\textwidth}
		\includegraphics[width=1\textwidth]{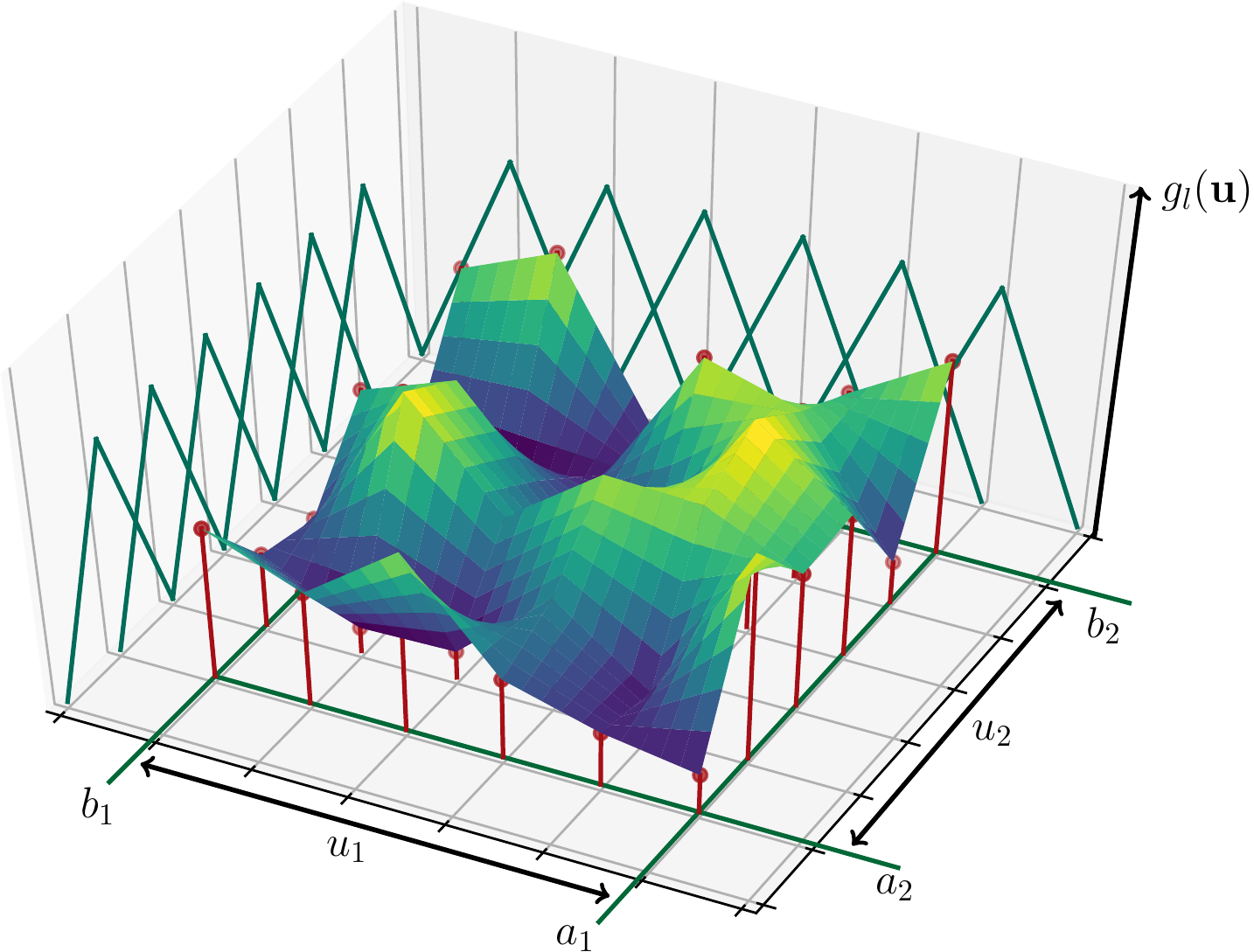}
		\caption{Linear B-spline basis functions}\label{fig:Ng1}
	\end{subfigure}
	\hfill
	\begin{subfigure}[t]{0.45\textwidth}
		\includegraphics[width=1\textwidth]{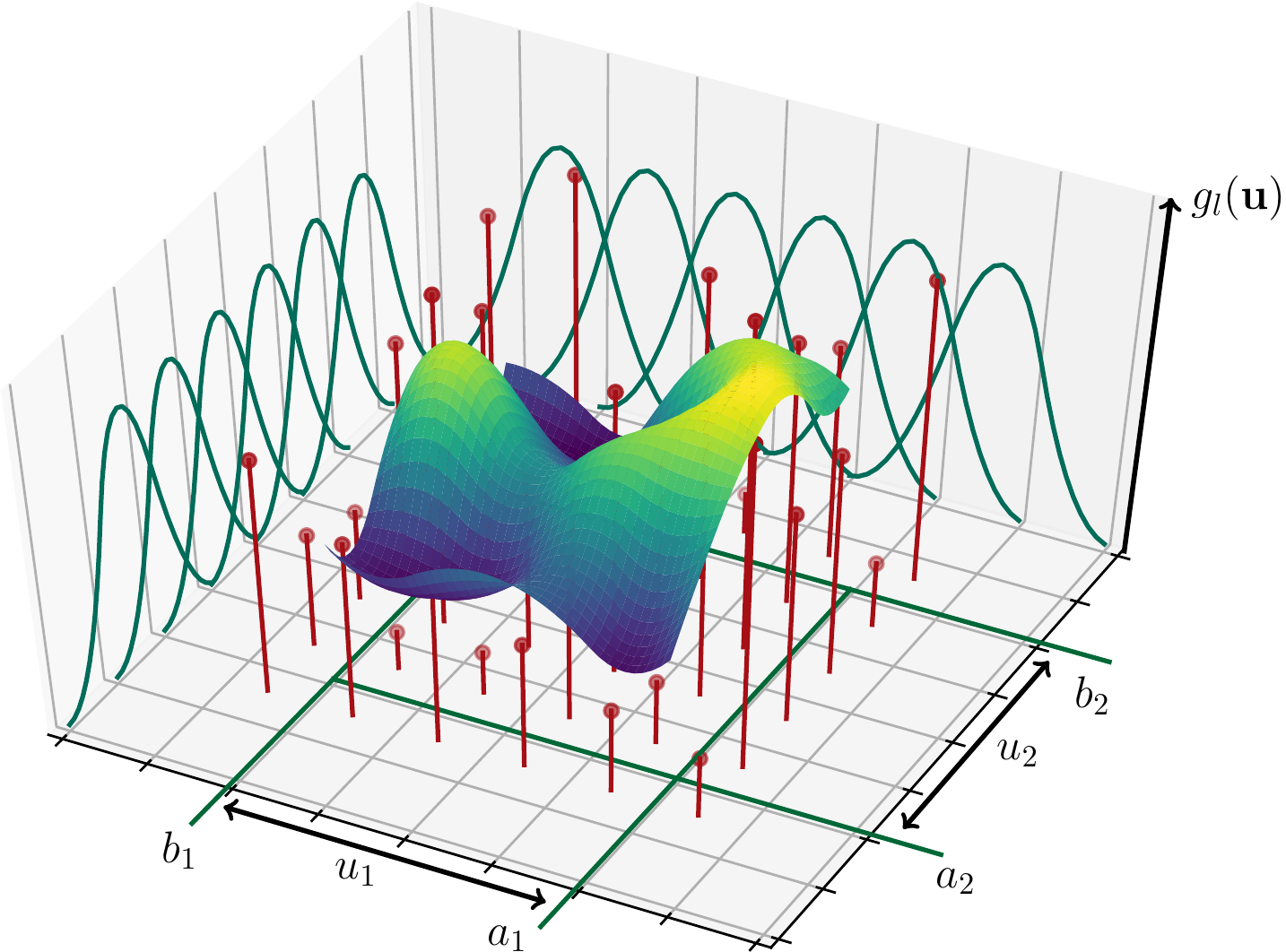}
		\caption{Quadratic B-spline basis functions}\label{fig:Ng2}
	\end{subfigure}
	\caption{Examples of our continuous convolution kernel for B-spline basis degrees (a) $m=1$  and (b) $m=2$  for kernel dimensionality $d=2$. The heights of the red dots are the trainable parameters for a single input feature map. They are multiplied by the elements of the B-spline tensor product basis before influencing the kernel value.}\label{fig:filter_img}
\end{figure*}

\subsection{Main concept}

Our convolution operator aggregates node features in local neighborhoods weighted by a trainable, continuous kernel function.
The node features $\mathbf{f}(i)$ represent features on an irregular geometric structure, whose spatial relations are locally defined by the pseudo-coordinates in $\mathbf{U}$.
Therefore, when locally aggregating feature values in a node's neighborhood, the content of $\mathbf{U}$ is used to determine \emph{how} the features are aggregated and the content of $\mathbf{f}(i)$ defines \emph{what} is aggregated.
We argue that common inputs for geometric deep learning tasks can be mapped to this model while preserving relevant information:
\begin{itemize}
	\setlength\itemsep{0.1em}
	\item For {\bf graphs}, $\mathcal{V}$ and $\mathcal{E}$ are given and $\mathbf{U}$ can contain edge weights or, for example, features like the node degree of the target nodes.
	\item For {\bf discrete manifolds}, $\mathcal{V}$ contains points of the discrete manifold, $\mathcal{E}$ represents connectivity in local Euclidean neighborhoods and $\mathbf{U}$ can contain local relational information like polar, spherical or Cartesian coordinates of the target point in respect to the origin point for each edge.
\end{itemize}
We state no restriction on the values of $\mathbf{U}$, except being element of a fixed interval range.
Therefore, meshes, for example, can be either interpreted as embedded three-dimensional graphs or as two-dimensional manifolds, using local Euclidean neighborhood representations like obtained by the work of Boscaini \etal~\cite{Boscaini2016}.
Also, either polar/spherical coordinates or Cartesian coordinates can be used, as shown in Figure~\ref{fig:pseudo}.
Independent from the type of coordinates stored in $\mathbf{U}$, our trainable, continuous kernel function, which we define in the following section, maps each $\mathbf{u}(i,j)$ to a scalar that is used as a weight for feature aggregation.

\subsection{Convolution operator}\label{sec:conv_operator}

We begin with the definition of a continuous kernel function using B-spline bases, which is parametrized by a constant number of trainable control values.
The local support property of B-spline basis functions~\cite{Piegl1997}, which states that basis functions evaluate to zero for all inputs outside of a known interval, proves to be advantageous for efficient computation and scalability.

Figure~\ref{fig:filter_img} visualizes the following kernel construction method for differing B-spline basis degree $m$.
We introduce a trainable parameter \mbox{$w_{\mathbf{p},l}\in \mathbf{W}$} for each element $\mathbf{p}$ from the Cartesian product \mbox{$ \mathcal{P} = {(N_{1,i}^m)}_i \times \cdots \times {(N_{d,i}^m)}_i$} of the B-spline bases and each of the $M_{\textrm{in}}$ input feature maps, indexed by $l$.
This results in $K = M_{\textrm{in}} \cdot \prod_{i=1}^{d} k_i$ trainable parameters.

We define our continuous convolution kernel as functions \mbox{$g_l:[a_1,b_1] \times \cdots \times [a_d,b_d] \rightarrow \mathbb{R}$} with
\begin{equation}
\label{eq:convolution_kernel}
g_l(\mathbf{u}) =  \sum_{\mathbf{p} \in \mathcal{P}} w_{\mathbf{p},l} \cdot B_{\mathbf{p}}(\mathbf{u}) \textrm{,}
\end{equation}
with $B_{\mathbf{p}}$ being the product of the basis functions in $\mathbf{p}$:
\begin{equation}
B_{\mathbf{p}}(\mathbf{u}) = \prod_{i=1}^{d}  N_{i,p_i}^m(u_i)\textrm{.}
\end{equation}
One way to interpret this kernel is to see the trainable parameters $w_{\mathbf{p},l}$ as control values for the height of a $d+1$-dimensional B-spline surface, from which a weight is sampled for each neighboring point $j$, depending on $\mathbf{u}(i,j)$.
However, in contrast to traditional $(d+1)$-dimensional B-spline approximation, we only have one-dimensional control points and approximate functions \mbox{$g_l:[a_1,b_1]\times \cdots \times [a_d,b_d] \rightarrow \mathbb{R}$} instead of curves.
The definition range of $g_l$ is the interval in which the partition of unity property of the B-spline bases holds~\cite{Piegl1997}.
Therefore, $a_i$ and $b_i$ depend on B-spline degree $m$ and kernel size $(k_1, \ldots, k_d)$.
We scale the spatial relation vectors $\mathbf{u}(i,j)$ to exactly match this interval, \cf~Figure~\ref{fig:filter_img}.

Given our kernel functions $\mathbf{g}=(g_1, \ldots ,g_{M_{\textrm{in}}})$ and input node features $\mathbf{f}$, we define our spatial convolution operator for a node $i$ as
\begin{equation}
\label{eq:convolution_operator}
(\mathbf{f} \star \mathbf{g})(i) = \frac{1}{|\mathcal{N}(i)|} \sum_{l = 1}^{M_{\textrm{in}}} \sum_{j \in \mathcal{N}(i)}  f_l(j) \cdot g_l(\mathbf{u}(i,j)).
\end{equation}
Similar to traditional CNNs, the convolution operator can be used as a module in a deep neural network architecture, which we do in our SplineCNNs.
To this end, the operator is applied $M_{\textrm{out}}$ times on the same input data with different trainable parameters, to obtain a convolutional layer that produces $M_{\textrm{out}}$ output feature maps. It should be highlighted that, in contrast to self-attention methods, we train an individual set of weights for each combination of input and output feature map.

\paragraph{Local support.}

Due to the local support property of B-splines, $B_{\mathbf{p}}\neq 0$ only holds true for $s:={(m+1)}^d$ of the $K$ different vectors $\mathbf{p}\in \mathcal{P}$.
Therefore, $g_l(\mathbf{u})$ only depends on $M_{\textrm{in}}\cdot s$ of the $M_{\textrm{in}}\cdot K$ trainable parameters for each neighbor $j$, where $s$, $d$ and $m$ are constant and usually small.
In addition, for each pair of nodes $(i,j)\in \mathcal{E}$, the vectors \mbox{$\mathbf{p} \in \mathcal{P}$} with $B_{\mathbf{p}}\neq 0$, which we denote as $\mathcal{P}(\mathbf{u}(i,j))$, can be found in constant time, given constant $m$ and $d$.

\begin{figure}
	\includegraphics[width=1\linewidth]{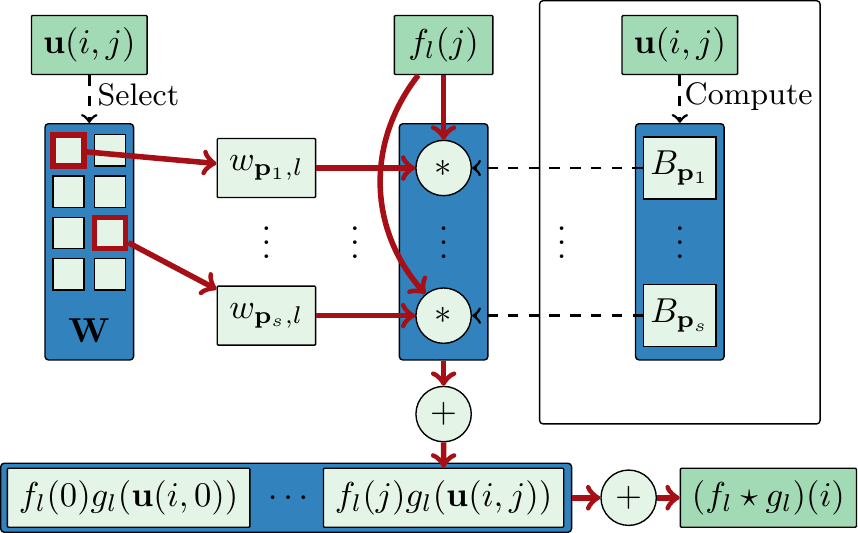}
	\caption{Forward computation scheme of the proposed convolution operation. During the backward step of the backpropagation algorithm, the gradient flows along the inverted solid arrows, reaching inputs from $\mathbf{W}$ and $f_l(i)$.}\label{fig:gradient_flow}
\end{figure}

This allows for an alternative representation of the inner sums of our convolution operation, \cf~Equation~\ref{eq:convolution_operator}, as
\begin{equation}
\label{eq:reformulation}
(f_l \star g_l)(i) = \sum_{\substack{j \in \mathcal{N}(i) \\ \mathbf{p} \in \mathcal{P}(\mathbf{u}(i,j))}} f_l(j) \cdot w_{\mathbf{p},l} \cdot B_{\mathbf{p}}(\mathbf{u}(i,j))  \textrm{.}
\end{equation}
and $K$ can be replaced by $s$ in the time complexity of the operation. Also, $B_{\mathbf{p}}(\mathbf{u}(i,j))$ does not depend on $l$ and can therefore be computed once for all input features.
Figure~\ref{fig:gradient_flow} shows a scheme of the computation. The gradient flow for the backward pass can also be derived by following the solid arrows backwards.

\paragraph{Closed B-splines.}

Depending on the type of coordinate in vectors $\mathbf{u}$, we use closed B-spline approximation in some dimensions.
One frequently occurring example of such a situation is when $\mathbf{u}$ contains angle attributes of polar coordinates.
Using closed B-spline approximation in the angle dimension naturally enforces the angle $0$ to be evaluated to the same weight as the angle $2\pi$ or, for higher $m$, the kernel function to be continuously differentiable at those points.

The proposed kernels can easily be modified so that they use closed approximation in an arbitrary subset of the $d$ dimensions, by mapping different $\mathbf{p} \in \mathcal{P}$ to the same trainable control value $w_{\mathbf{p},l}$.
This leads to a reduction of trainable parameters and B-spline basis functions.
Referring to Figure~\ref{fig:filter_img}, this approach can be interpreted as periodic repetition of the function surface along the corresponding axis.

\paragraph{Root nodes.}

Up to now, we did not consider the node $i$ of neighborhood $\mathcal{N}(i)$ in our convolution operator.
It is not aggregated together with all $j \in \mathcal{N}(i)$, like it would be the case in traditional CNNs.
If Cartesian coordinates are used, we can simply define $\mathcal{N}(i)$ to include $i$.
However, when using polar/spherical pseudo-coordinates, problems arise since the point with zero radius is not well defined.
Therefore, we introduce an additional trainable parameter for each feature of the root node and add the product of this parameter and the corresponding feature to the result.

\paragraph{Relation to traditional CNNs.}

Except for a normalization factor, our spline-based convolution operator is a generalization of the traditional convolutional layer in CNNs with \emph{odd} filter size in each dimension.
For example, if we assume to have a two-dimensional grid-graph with diagonal, horizontal and vertical edges to be the input, B-spline degree $m=1$, kernel size $(3, 3)$, and the vectors $\mathbf{u}$ to contain Cartesian relations between adjacent nodes, then our convolution operator is equivalent to a discrete convolution of an image with a kernel of size $3\times3$.
This also holds for larger discrete kernels if the neighborhoods of the grid-graph are modified accordingly.

\section{GPGPU algorithm}

For the spline-based convolutional layer defined in the last section, we introduce a GPU algorithm which allows efficient training and inference with SplineCNNs.
For simplicity, we use a tensor indexing notation with, \eg, $\mathbf{A}[x,y,z]$ describing the element at position $(x,y,z)$ of a tensor $\mathbf{A}$ with rank three.
Our forward operation of our convolution operator is outlined in Algorithm~\ref{alg:forward}.

\begin{algorithm}[tbp]
	\caption{Geometric convolution with B-spline kernels}
	\begin{algorithmic}\label{alg:forward}
		\STATE{} \textbf{Input}:
		\STATE{} $N$: Number of nodes
		\STATE{} $M_{\textrm{in}}$: Number of input features per node
		\STATE{} $M_{\textrm{out}}$: Number of output features per node
		\STATE{} $s = {(m+1)}^d$: Number of non-zero $B_{\mathbf{p}}$ for one edge
		\STATE{} $\mathbf{W} \in \mathbb{R}^{K \times M_{\textrm{in}} \times M_{\textrm{out}} }$: Trainable weights
		\STATE{} $\mathbf{B} \in \mathbb{R}^{E\times s}$: Basis products of $s$ weights for each edge
		\STATE{} $\mathbf{P} \in \mathbb{N}^{E\times s}$: Indices of $s$ weights in $\mathbf{W}$ for each edge
		\STATE{} $\mathbf{F}_{\textrm{in}} \in \mathbb{R}^{N\times M_{\textrm{in}}}$: Input features for each node
		\STATE{} \textbf{Output}:
		\STATE{} $\mathbf{F}_{\textrm{out}}\in \mathbb{R}^{N\times M_{\textrm{out}}}$: Output features for each node
		\STATE{} -------------------------------------------------------------------- %chktex 8
		\STATE{} Gather $\mathbf{F}^E_{\textrm{in}}$ from $\mathbf{F}_{\textrm{in}}$ based on target nodes of edges
		\STATE{} {\bf Parallelize} over $e \in \{1,\ldots,E\}$, $o \in \{1,\ldots,M_{\textrm{out}}\}$:
		\bindent%
		\STATE{} $r \leftarrow 0$
		\FOR{each $i \in \{1,\ldots,M_{\textrm{in}}\}$}
		\FOR{each $p \in \{1,\ldots,s\}$}
		\STATE{} $w \leftarrow \mathbf{W}[\mathbf{P}[e,p],i,o]$
		\STATE{} $r \leftarrow r + ( \mathbf{F}^E_{\textrm{in}}[e,i] \cdot w \cdot \mathbf{B}[e,p])$
		\ENDFOR%
		\ENDFOR%
		\STATE{} $\mathbf{F}^E_{\textrm{out}}[e,o] \leftarrow r$
		\eindent%
		\STATE{} Scatter-add $\mathbf{F}^E_{\textrm{out}}$ to $\mathbf{F}_{\textrm{out}}$ based on origin nodes of edges
		\STATE{} Return $\mathbf{F}_{\textrm{out}}$
	\end{algorithmic}
\end{algorithm}

We achieve parallelization over the edges $\mathcal{E}$ by first gathering edge-wise input features $\mathbf{F}^E_{\textrm{in}} \in \mathbb{R}^{E\times M_{\textrm{in}}}$ from the input matrix $\mathbf{F}_{\textrm{in}}\in \mathbb{R}^{N\times M_{\textrm{in}}}$, using the target node of each edge as index.
Then, we compute edge-wise output features $\mathbf{F}^E_{\textrm{out}} \in \mathbb{R}^{E\times M_{\textrm{out}}}$, as shown in Figure~\ref{fig:gradient_flow}, before scatter-adding them back to node-wise features $\mathbf{F}_{\textrm{out}}\in \mathbb{R}^{N\times M_{\textrm{out}}}$, performing the actual neighborhood aggregation.
Our algorithm has a parallel time complexity of $\mathcal{O}(s \cdot M_{\textrm{in}})$, with small $s$, using $\mathcal{O}(E \cdot M_{\textrm{out}})$ processors, assuming that scatter-add is a parallel operation with constant time complexity.

\paragraph{Computing B-spline bases.}

We achieve independence from the number of trainable weights by computing matrices $\mathbf{P} \in \mathbb{N}^{E\times s}$ and $\mathbf{B} \in \mathbb{R}^{E\times s}$.
$\mathbf{P}$ contains the indices of parameters with $B_{\mathbf{p}}\neq 0$ while $\mathbf{B}$ contains the basis products $B_{\mathbf{p}}$ for these parameters.
$\mathbf{B}$ and $\mathbf{P}$ can be preprocessed for a given graph structure or can be computed directly in the kernel.
For the GPU evaluation of the basis functions required for $\mathbf{B}$ we use explicit low-degree polynomial formulations of those functions for each $m$.
For further details we refer to our PyTorch implementation, which is available on GitHub.

\paragraph{Mini-batch handling.}

For batch learning, parallelization over a mini-batch can be achieved by creating sparse block diagonal matrices out of all $\mathbf{U}$ of one batch and concatenating matrices $\mathbf{F}_{\textrm{in}}$ in the node dimension.
For matrices $\mathbf{F}^E_{\textrm{in}}$, $\mathbf{B}$ and $\mathbf{P}$, this results in example-wise concatenation in the edge dimension.
Note that this composition allows differing number of nodes and edges over examples in one batch without introducing redundant computational overhead.

\section{Results}\label{sec:results}

We perform experiments with different SplineCNN architectures on three distinct tasks from the fields of image graph classification (Section~\ref{sec:image_graph}), graph node classification (Section~\ref{sec:graph_node}) and shape correspondence on meshes (Section~\ref{sec:shape_corr}).
For each of the tasks, we create a SplineCNN using the spline-based convolution operator which we denote as SConv($\mathbf{k},M_{\textrm{in}},M_{\textrm{out}}$) for a convolutional layer with kernel size $\mathbf{k}$, $M_{\textrm{in}}$ input feature maps and $M_{\textrm{out}}$ output feature maps. % chktex 36
In addition, we denote fully connected layers as FC($o$), with $o$ as number of output neurons. % chktex 36

\subsection{Image graph classification}\label{sec:image_graph}

\begin{figure}[t]
	\begin{subfigure}[b]{0.49\linewidth}
		\includegraphics[width=1\linewidth]{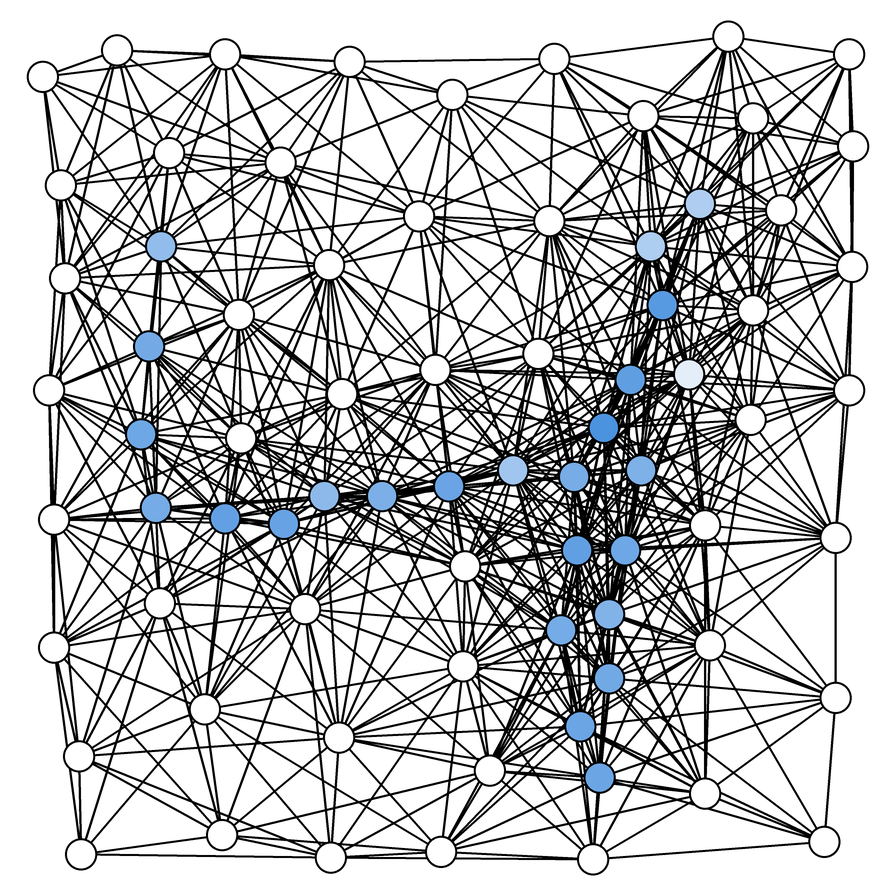}
		\caption{MNIST superpixels example}\label{fig:mnist_superpixel_graph}
	\end{subfigure}
	\begin{subfigure}[b]{0.49\linewidth}
		\includegraphics[width=1\linewidth]{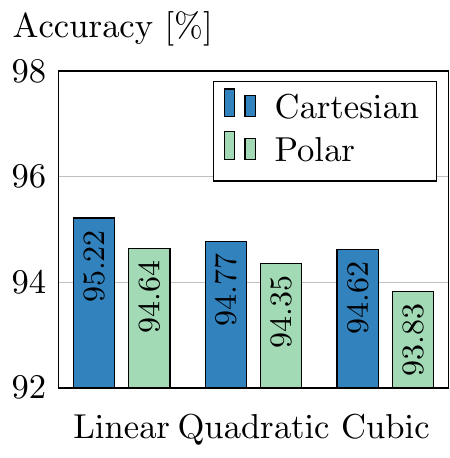}
		\caption{Classification accuracy}\label{fig:mnist_spline_acc}
	\end{subfigure}
	\caption{MNIST 75 superpixels (a) example and (b) classification accuracy of SplineCNN using varying pseudo-coordinates and B-spline base degrees.}\label{fig:mnist_images}
\end{figure}

For validation on two-dimensional regular and irregular structured input data, we apply our method on the widely-known MNIST dataset~\cite{Lecun1998} of 60,000 training and 10,000 test images containing grayscale, handwritten digits from $10$ different classes.
We conduct two different experiments on MNIST. %chktex 13
For both experiments, we strictly follow the experimental setup of Defferrard \etal~and Monti \etal~\cite{Defferrard2016, Monti2016} to provide comparability.
For the first experiment, the MNIST images are represented as a set of \emph{equal} grid graphs, where each node corresponds to one pixel in the original image, resulting in grids of size $28 \times 28$ with $N = 28^2 = 784$ nodes.
For the second experiment, the MNIST superpixel dataset of Monti \etal~\cite{Monti2016} is used, where each image is represented as an embedded graph of 75 nodes defining the centroids of superpixels, \cf~Figure~\ref{fig:mnist_superpixel_graph}, with each graph having \emph{different} node positions and connectivities.
This experiment is an ideal choice to validate the capabilities of our approach on irregular structured, image-based data.

\paragraph{Pooling.}

Our SplineCNN architectures use a pooling operator based on the Graclus method~\cite{Dhillon2007, Defferrard2016}.
The pooling operation is able to obtain a coarsened graph by deriving a clustering on the graph nodes, aggregating nodes in one cluster and computing new pseudo-coordinates for each of those new nodes.
We denote a max-pooling layer using this algorithm with MaxP($c$), with $c$ being the cluster size (and approximate downscaling factor). % chktex 36

\begin{table}
	\begin{center}
		\begin{tabular}{lccc}
			\toprule
			Dataset & LeNet5~\cite{Lecun1998} & MoNet~\cite{Monti2016} & \textbf{SplineCNN} \\
			\midrule
			Grid & \textbf{99.33\%} & 99.19\% & 99.22\% \\
			Superpixels & -- & 91.11\% & \textbf{95.22\%} \\ % chktex 8
			\bottomrule
		\end{tabular}
	\end{center}
	\caption{Classification accuracy on different representations of the MNIST dataset (grid and superpixel) for a classical CNN (LeNet5), MoNet and our SplineCNN approach.}\label{tab:mnist_results}
\end{table}

\paragraph{Architectures and parameters.}

For the grid graph experiments, Cartesian coordinates and a B-spline basis degree of $m=1$ are used to reach equivalence to the traditional convolution operator in CNNs, \cf~Section~\ref{sec:conv_operator}.
In contrast, we compare all configurations of $m$ and possible pseudo-coordinates against each other on the superpixel dataset.

For classification on the grid data, we make use of a LeNet5-like network architecture~\cite{Lecun1998}: SConv($(5,5)$,$1$,$32$) $\rightarrow$ MaxP(4) $\rightarrow$ SConv(($5,5$),$32$,$64$) $\rightarrow$ MaxP(4) $\rightarrow$ FC($512$) $\rightarrow$ FC($10$). % chktex 36
The initial learning rate was chosen as $10^{-3}$ and dropout probability as $0.5$.
Note that we used neighborhoods of size $5 \times 5$ from the grid graph, to mirror the LeNet5 architecture with its $5 \times 5$ filters.

The superpixel dataset is evaluated using the SplineCNN architecture SConv($(k_1,k_2)$,$1$,$32$) $\rightarrow$ MaxP(4) $\rightarrow$ SConv(($k_1,k_2$),$32$,$64$) $\rightarrow$ MaxP(4) $\rightarrow$ AvgP $\rightarrow$  FC($128$) $\rightarrow$  FC($10$), where AvgP denotes a layer that averages features in the node dimension. % chktex 36
We use the \emph{Exponential Linear Unit (ELU)} as non-linearity after each SConv layer and the first FC layer.
For Cartesian coordinates, we choose the kernel size to be $k_1 = k_2 =4+m$ and for polar coordinates $k_1=1+m$ and $k_2=8$.
Training was done for 20 epochs with a batch size of $64$, initial learning rate $0.01$ and dropout probability $0.5$.
Both networks were trained for 30 epochs using the Adam method~\cite{Kingma2015}.

\paragraph{Discussion.}

All results of the MNIST experiments are shown in Table~\ref{tab:mnist_results} and Figure~\ref{fig:mnist_spline_acc}.
The grid graph experiment results in approximately the same accuracy as LeNet5 and the MoNet method.
For the superpixel dataset, we improve previous results by $4.11$ percentage points in accuracy.
Since we are using a similar architecture and the same input data as MoNet, the better results are an indication that our operator is able to capture more relevant information in the structure of the input.
This can be explained by the fact that, in contrast to the MoNet kernels, our kernel function has individual trainable weights for each combination of input and output feature maps, just like the filters in traditional CNNs.

Results for different configurations are shown in Figure~\ref{fig:mnist_spline_acc}.
We only notice small differences in accuracy for varying $m$ and pseudo-coordinates.
However, lower $m$ and using Cartesian coordinates performs slightly better than the other configurations.

\begin{figure}
	\begin{subfigure}[t]{\linewidth}
		\includegraphics[width=1\linewidth]{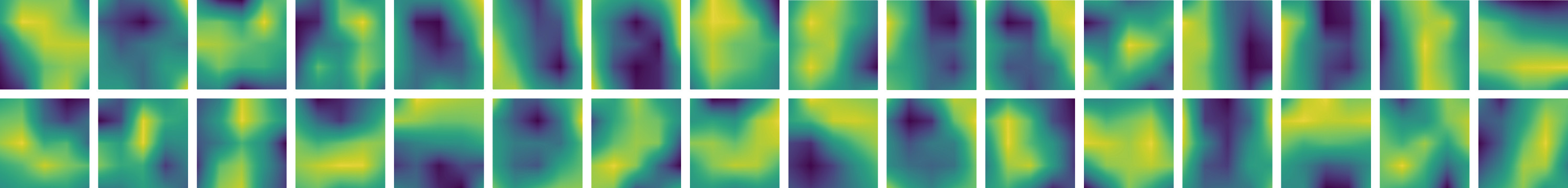}
		\caption{MNIST grid experiment}
	\end{subfigure}\\[1.5ex] %Some more spacing between caption and next figure
	\begin{subfigure}[t]{\linewidth}
		\includegraphics[width=1\linewidth]{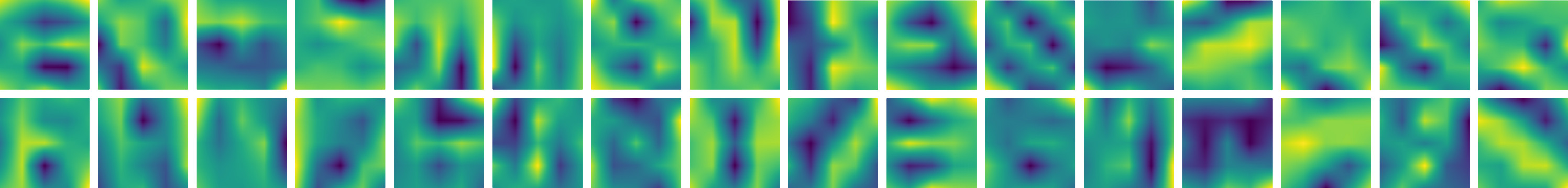}
		\caption{MNIST superpixel experiment}
	\end{subfigure}
	\caption{Visualizations of the 32 kernels from the first spline-based convolutional layers, trained on the MNIST (a) grid and (b) superpixels datasets, with kernel size $(5, 5)$ and B-spline base degree $m=1$.}\label{fig:mnist_kernel}
\end{figure}

In addition, we visualized the $32$ learned kernels of the first SConv layers from the grid and superpixel experiments in Figure~\ref{fig:mnist_kernel}.
It can be observed that edge detecting patterns are learned in both approaches, whether being trained on regular or irregular structured data.

\subsection{Graph node classification}\label{sec:graph_node}

\begin{table}
	\begin{center}
		\resizebox{\linewidth}{!}{%
			\begin{tabular}{cccc}
				\toprule
				ChebNet~\cite{Defferrard2016} & GCN~\cite{Kipf2017} & CayleyNet~\cite{Levie2017} & \textbf{SplineCNN} \\
				\midrule
				87.12 $\pm$ 0.60 & 87.17 $\pm$ 0.58 & 87.90 $\pm$ 0.66 & \textbf{89.48 $\pm$ 0.31} \\
				\bottomrule
			\end{tabular}}
		\end{center}
		\caption{Graph node classification on the Cora dataset for different learning methods (ChebNet, GCN, CayleyNet and SplineCNN). The presented accuracy means and standard deviations are computed over 100 experiments, where for each experiment the network was trained for 200 epochs.}\label{tab:cora_results}
	\end{table}
	
	As second experiment, we address the problem of graph node classification using the Cora citation graph~\cite{Sen2008}.
	We validate that our method also performs strongly on datasets, where no Euclidean relations are given.
	Cora consists of 2,708 nodes and 5,429 undirected unweighted edges, representing scientific publications and citation links respectively.
	Each document is represented individually by a 1,433 dimensional sparse binary bag-of-words feature vector and is labeled to exactly one out of 7 classes.
	Similar to the experimental setup in Levi \etal~\cite{Levie2017}, we split the dataset into 1,708 nodes for training and 500 nodes for testing, to simulate labeled and unlabeled information.
	
	\paragraph{Architecture and parameters.}
	
	We use a SplineCNN similar to the network architecture introduced \mbox{in~\cite{Levie2017, Kipf2017, Monti2016}}: SConv($(2)$,$1433$,$16$) $\rightarrow$ SConv($(2)$,$16$,$7$), with ELU activation after the first SConv layer and $m=1$. % chktex 36
	For pseudo-coordinates, we choose the globally normalized degree of the target nodes $\textbf{u}(i,j) = (\deg(j)/\max_{v\in\mathcal{V}} \deg(v))$, leading to filtering based on the number of cites of neighboring publications.
	Training was done using the Adam optimization method~\cite{Kingma2015} for 200 epochs with learning rate $0.01$, dropout probability $0.5$ and L2 regularization $0.005$.
	As loss function, the cross entropy between the network's softmax output and a one-hot target distribution was used.
	
	\paragraph{Discussion.}
	
	Results of our and related methods are shown in Table~\ref{tab:cora_results} and report the mean classification accuracy averaged over 100 experiments.
	It can be seen that \mbox{SplineCNNs} improve the state-of-the-art in this experiment by approximately $1.58$ percentage points.
	We contribute this improvement to the filtering based on $\mathbf{u}$, which contains node degrees as additional information to learn more complex kernel functions.
	This indicates that SplineCNNs can be successfully applied to irregular but non-geometric data and that they are able to improve previous results in this domain.
	
	\subsection{Shape correspondence}\label{sec:shape_corr}
	
	\begin{figure*}[t]
		\begin{subfigure}[b]{0.32\textwidth}
			\begin{center}
				\includegraphics[width=1\linewidth]{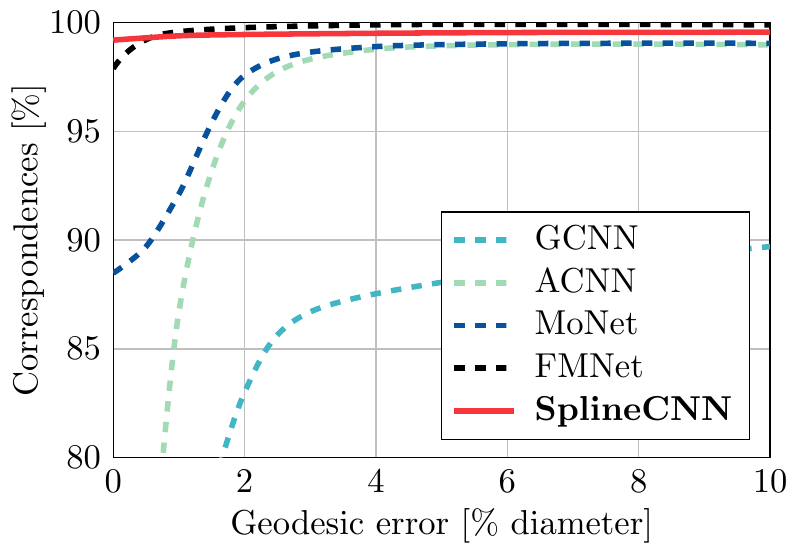}
			\end{center}
			\vspace{-10pt}
			\caption{Results of SplineCNN and other methods}\label{fig:geodesic_error_other}
		\end{subfigure}
		\begin{subfigure}[b]{0.32\textwidth}
			\begin{center}
				\includegraphics[width=1\linewidth]{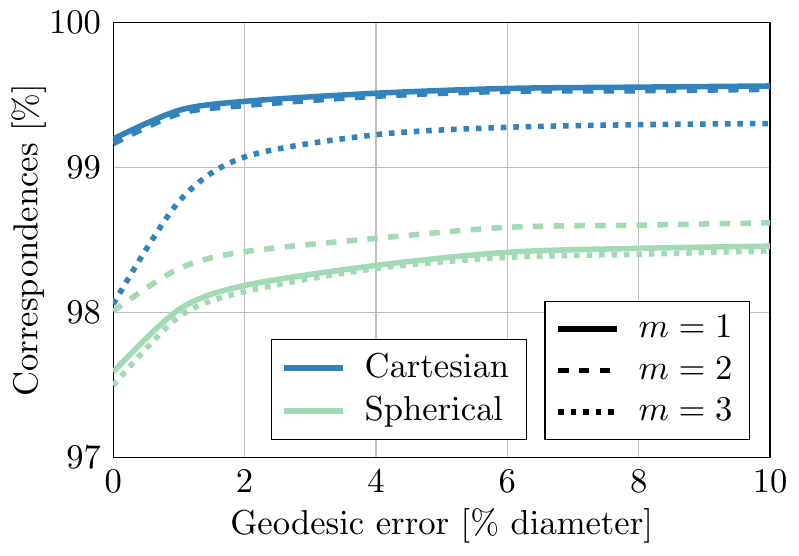}
			\end{center}
			\vspace{-10pt}
			\caption{Results for different SplineCNNs}\label{fig:geodesic_error_our}
		\end{subfigure}
		\begin{subfigure}[b]{0.34\textwidth}
			\begin{center}
				\includegraphics[width=0.9\linewidth]{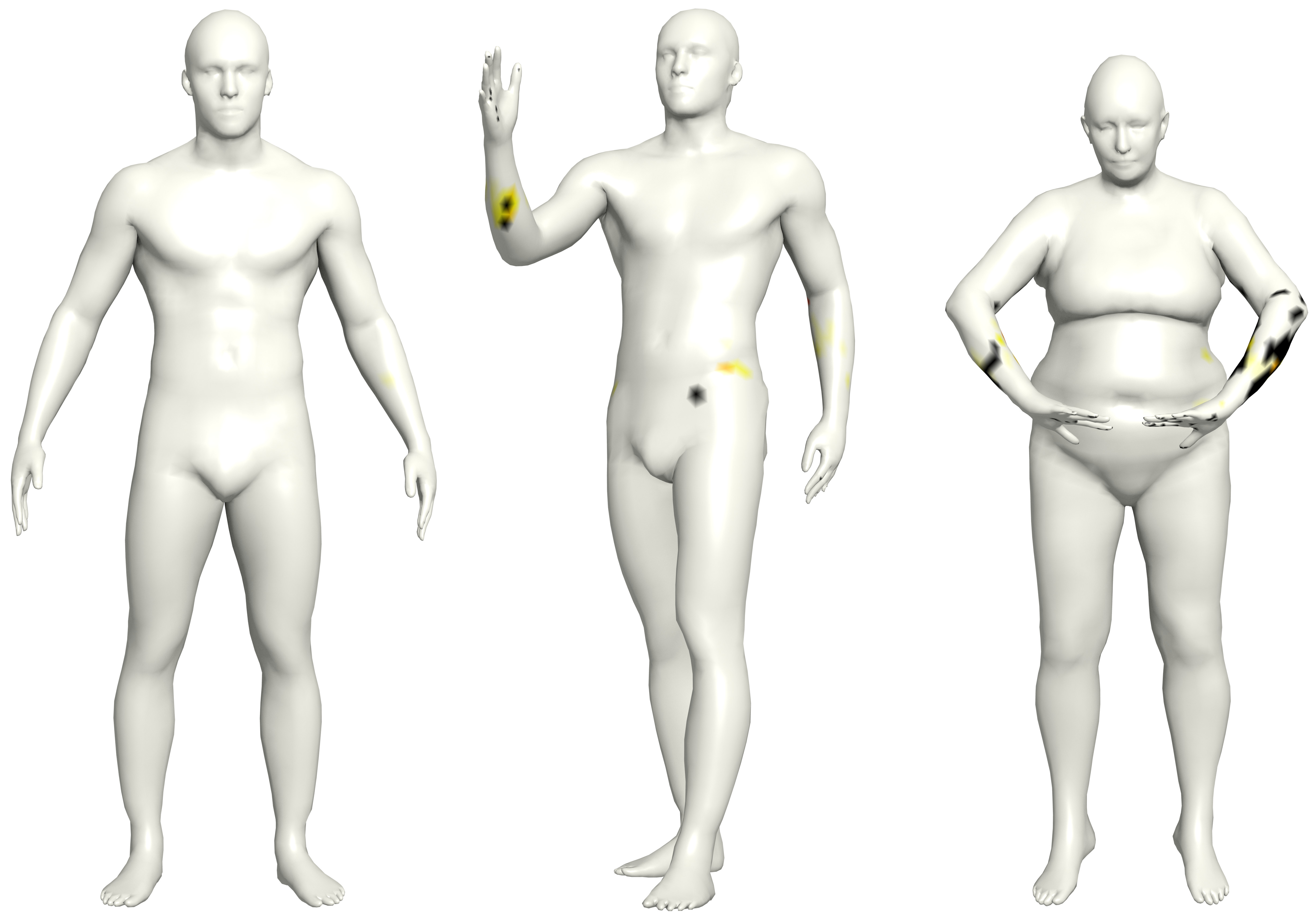}
				\includegraphics[width=0.08\linewidth]{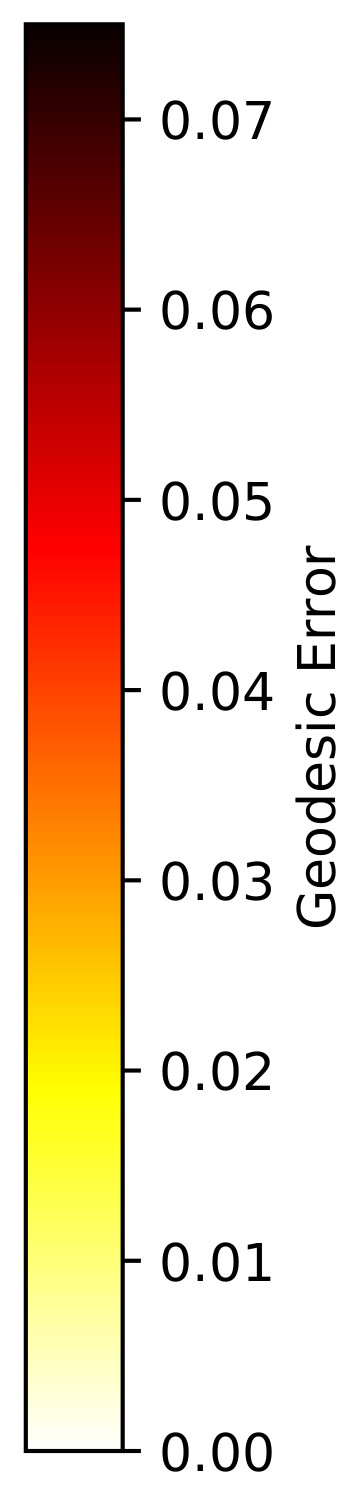}
			\end{center}
			\vspace{-10pt}
			\caption{Geodesic error of test examples}\label{fig:geodesic_error_img}
		\end{subfigure}
		\caption{%
			Geodesic error plots of the shape correspondence experiments with (a) SplineCNN and related approaches and (b) different SplineCNN experiments. The x-axis displays the geodesic distance in \% of diameter and the y-axis the percentage of correspondences that lie within a given geodesic radius around the correct node.
			Our SplineCNN achieves the highest accuracy for low geodesic error and significantly outperforms other general approaches like MoNet, GCNN and ACNN. % chktex 13
			In Figure (c), three examples of the FAUST test dataset with geodesic errors of SplineCNN predictions for each node are presented. We show the best (left), the median (middle) and worst (right) test example, sorted by average geodesic error.
		}\label{fig:geodesic_error}
	\end{figure*}
	
	As our last and largest experiment, we validate our method on a collection of three-dimensional meshes solving the task of shape correspondence similar to~\cite{Monti2016, Boscaini2016, Masci2015, Litany2017}.
	Shape correspondence refers to the task of labeling each node of a given shape to the corresponding node of a reference shape~\cite{Masci2015}.
	We use the FAUST dataset~\cite{Bogo2014}, containing 10 scanned human shapes in 10 different poses, resulting in a total of 100 non-watertight meshes with 6,890 nodes each.
	The first 80 subjects in FAUST were used for training and the remaining 20 subjects for testing, following the dataset splits introduced in~\cite{Monti2016}.
	Ground truth correspondence of FAUST meshes are given implicitly, where nodes are sorted in the exact same order for every example.
	Correspondence quality is measured according to the Princeton benchmark protocol~\cite{Kim2011}, counting the percentage of derived correspondences that lie within a geodesic radius $r$ around the correct node.
	
	In contrast to similar approaches, \eg~\cite{Monti2016, Boscaini2016, Masci2015, Litany2017}, we go without handcrafted feature descriptors as inputs, like the local histogram of normal vectors known as SHOT descriptors~\cite{Tombari2010}, and force the network to learn from the geometry (\ie~spatial relations encoded in $\mathbf{U})$ itself.
	Therefore, input features are trivially given by $\mathbf{1} \in \mathbb{R}^{N \times 1}$.
	Also, we validate our method on three-dimensional meshes as inputs instead of generating two-dimensional geodesic patches for each node.
	These simplifications reduce the computation time and memory consumption that are required to preprocess the data by a wide margin, making training and inference completely end-to-end and very efficient.
	
	\paragraph{Architecture and parameters.}
	
	We apply a Spline\-CNN architecture with 6 convolutional layers: SConv($(k_1, k_2, k_3)$,$1$,$32$) $\rightarrow$ SConv($(k_1, k_2, k_3)$,$32$,$64$) $\rightarrow$ $4\times$ SConv($(k_1, k_2, k_3)$,$64$,$64$) $\rightarrow$ Lin($256$) $\rightarrow$ Lin($6890$), where Lin($o$) denotes a \mbox{$1 \times 1$} convolutional layer to $o$ output features per node. % chktex 36
	As non-linear activation function, ELU is used after each SConv and the first Lin layer.
	For Cartesian coordinates we choose the kernel size to be \mbox{$k_1 = k_2 = k_3 = 4 + m$} and for polar coordinates \mbox{$k_1 = k_3 = 4 + m$} and \mbox{$k_2 = 8$}.
	We evaluate our method on multiple choices of $m = \{1, 2, 3\}$.
	Training was done for 100 epochs with a batch size of 1, initial learning rate $0.01$ and dropout probability $0.5$, using the Adam optimizer~\cite{Kingma2015} and cross entropy loss.
	
	\paragraph{Discussion.}
	
	Obtained accuracies for different geodesic errors are plotted in Figure~\ref{fig:geodesic_error}.
	The results for different SplineCNN parameters match the observations from before, where only small differences could be seen but using Cartesian coordinates and small B-spline degrees seemed to be slightly better.
	Our SplineCNN outperforms all other approaches with $99.20 \%$ of predictions on the test set having \emph{zero} geodesic error.
	However, the global behavior over larger geodesic error bounds is slightly worse in comparison to FMNet~\cite{Litany2017}.
	In Figure~\ref{fig:geodesic_error_img} it can be seen that most nodes are classified correctly but that the few false classifications have a high geodesic error.
	We contribute this differences to the varying loss formulations.
	While we train against a one-hot binary vector using the cross entropy loss, FMNet trains using a specialized soft error loss, which is a more geometrically meaningful criterion that punishes geodesically far-away predictions stronger than predictions near the correct node~\cite{Litany2017}.
	However, it is worth highlighting that we do not use SHOT descriptors as input features, like all other approaches we compare against.
	Instead, we train only on the geometric structure of the meshes.
	
	\paragraph{Performance} We report an average forward step runtime of $0.043$ seconds for a single FAUST example processed by the suggested SplineCNN architecture ($k_1 = k_2 = k_3 = 5$, $m=1$) on a single NVIDIA GTX 1080 Ti.
	We train this network in approximately $40$ minutes.
	Regarding scalability, we are able to stack up to 160 SConv($(5,5,5)$,$64$,$64$) layers before running out of memory on the mentioned GPU, while the runtime scales linearly with the number of layers. However, for this task we do not observe significant improvement in accuracy when using deeper networks. % chktex 36
	
	\section{Conclusion}
	
	We introduced SplineCNN, a spline-based convolutional neural network with a novel trainable convolution operator, which learns on irregular structured, geometric input data.
	Our convolution filter operates in the spatial domain and aggregates local features, applying a trainable continuous kernel function parametrized by trainable B-spline control values.
	We showed that SplineCNN is able to improve state-of-the-art results in several benchmark tasks, including image graph classification, graph node classification and shape correspondence on meshes, while allowing very fast training and inference computation.
	To conclude, SplineCNN is the first architecture that allows deep end-to-end learning directly from geometric data while providing strong results. Due to missing preprocessing, this allows for even faster processing of data.
	
	In the future we plan to enhance SplineCNNs by concepts known from traditional CNNs, namely recurrent neurons for geometric, spatio-temporal data or dynamic graphs, and un-pooling layers to allow encoder-decoder or generative architectures.

\section*{Acknowledgments}
\noindent
This work has been supported by the \emph{German Research Association (DFG)}  within the Collaborative Research Center SFB~876, 
\emph{Providing Information by Resource-Constrained Analysis}, 
projects B2 and A6.
We also thank Pascal Libuschewski for proofreading and helpful advice.

{\small
\bibliographystyle{ieee}
\bibliography{egbib}
}

\end{document}